\documentclass{article}

\usepackage{arxiv}

\usepackage[utf8]{inputenc} 
\usepackage[T1]{fontenc}    
\usepackage{hyperref}       
\usepackage{url}            
\usepackage{booktabs}       
\usepackage{amsfonts}       
\usepackage{nicefrac}       
\usepackage{microtype}      
\usepackage{lipsum}
\usepackage{graphicx}
\graphicspath{ {./images/} }

\usepackage{cite}
\begin{document}

\title{Reconfigurable Cyber-Physical System for Lifestyle Video-Monitoring via Deep Learning}

\author{
    Daniel ~Deniz \\
    Computer Architecture and Technology \\
    CITIC, University of Granada\\
    Granada, Spain \\
    \texttt{danideniz@ugr.es} \\
\And
    Francisco ~Barranco \\
    Computer Architecture and Technology \\
    CITIC, University of Granada\\
    Granada, Spain \\
    \texttt{fbarranco@ugr.es} \\
\And
    Juan ~Isern \\
    Computer Architecture and Technology \\
    CITIC, University of Granada\\
    Granada, Spain \\
    \texttt{jisern@ugr.es} \\

\And
    Eduardo ~Ros \\
    Computer Architecture and Technology \\
    CITIC, University of Granada\\
    Granada, Spain \\
    \texttt{eros@ugr.es}
}

\maketitle

\begin{abstract}
Indoor monitoring of people at their homes has become a popular application in Smart Health.
With the advances in Machine Learning and hardware for embedded devices, new distributed approaches for Cyber-Physical Systems (CPSs) are enabled. Also, changing environments and need for cost reduction motivate novel reconfigurable CPS architectures. In this work, we propose an indoor monitoring reconfigurable CPS that uses embedded local nodes (Nvidia Jetson TX2). We embed Deep Learning architectures to address Human Action Recognition. Local processing at these nodes let us tackle some common issues: reduction of data bandwidth usage and preservation of privacy (no raw images are transmitted). Also real-time processing is facilitated since optimized nodes compute only its local video feed. Regarding the reconfiguration, a remote platform monitors CPS qualities and a Quality and Resource Management (QRM) tool sends commands to the CPS core to trigger its reconfiguration. Our proposal is an energy-aware system that triggers reconfiguration based on energy consumption for battery-powered nodes. Reconfiguration reduces up to 22\% the local nodes energy consumption extending the device operating time, preserving similar accuracy with respect to the alternative with no reconfiguration.

\end{abstract}


\section{Introduction}
\label{sec:introduction}

Fueled by the rapid evolution of smart integration of Systems-on-Chip (SoC), the Internet of the Things (IoT) and cloud computing, CPS are becoming omnipresent in many key application domains. CPS refers to a system that tightly integrates software computing modules, network components, and physical processes \cite{lee2008cyber}\cite{ding2019survey}. Modern CPS are becoming distributed architectures of highly-connected smart embedded systems that are increasingly autonomous \cite{al2018cyber}. CPS technologies are present in fields such as automotive industry with for example platoons of autonomous vehicles \cite{li2019design}, ensuring security in critical infrastructures for Smart-grid \cite{pal2016stream}, cloud computing \cite{chaari2016cyber}, or management systems for Smart-Health \cite{zhang2015health}.

\begin{figure}[t]
\begin{center}
\includegraphics[width=0.85\linewidth]{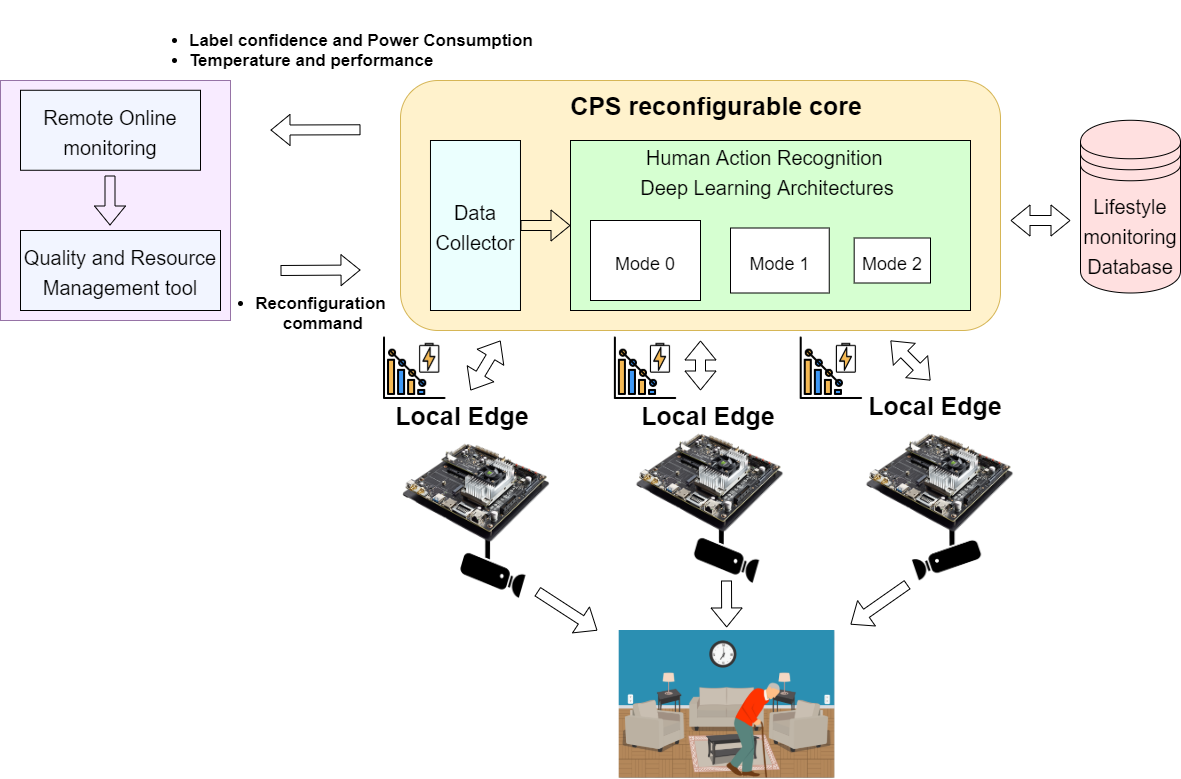}
\end{center}
    \caption{System overview: Multiple local edges perform lifestyle indoor action recognition. The reconfigurable CPS core gathers processed data and metric results from nodes, and sends it to the remote online monitoring tool. The logged data is analyzed by the quality and resource management tool that makes decisions about the reconfiguration of the system to achieve the overall system goal. In the diagram, it sends a reconfiguration command to the HAR module to select and re-deploy the fittest DL model in local edges. The goal is to maximize the working time of the battery-powered device. Monitored raw data and analyzed data are stored in the lifestyle monitoring database.}
\label{figure:overall}
\end{figure}

Given the changing environments common in many of the applications fields, CPSs are in need of different types of sensors, network topologies, or configurations. Novel approaches foster reconfiguration or adaptation of CPSs instead of re-deployment of hardware and sensors, in order to reduce cost and time \cite{bharad2014towards}. Reconfiguration enables modular and scalable run-time optimization to provide quality and on-line performance improvement, even across the different reconfigurations \cite{al2019fitoptivis}.

On-line monitoring techniques are a crucial component towards enabling the adaptation to the different reconfiguration profiles \cite{barrere2018cps}. Both reconfiguration profiles and resource requirements are optimized for the changing operational contexts. Finally, quality and resource management (QRM) tools actually make decisions about the optimal reconfiguration alternative driven by objectives such as energy efficiency, data bandwidth, performance, or cost \cite{hendriks2020interface}.

Visual information plays a crucial role for context awareness. Modern high-performance visual processing systems are computationally intensive and costly. Moreover, they require high data bandwidth for communication, and are subjected to long latencies due to the transmission delays \cite{you2017scaling}. On the opposite side, vision CPSs provide distributed solutions where most computing takes place at local edges that are smart embedded systems-on-chip in heterogeneous architectures \cite{barranco2014real}. These local edge nodes only transmit results that are gathered by the CPS core to address global actions. This distributed paradigm benefits from the reduction of bandwidth usage and power consumption.

A fundamental component of our lifestyle monitoring system is based on Human Activity Recognition (HAR). HAR systems collect data from human activity behavior and automatically classify this data with an action label \cite{lara2012survey}. This application has recently become very popular, with systems identifying human actions from diverse multi-modal sources and using different technologies \cite{huang2019tse}. Our objective in this work is the development of a lifestyle monitoring system to improve the well-being of people at their own homes, processing video feeds from different cameras. The system will recognize and gather the information from subjects at their homes. This data will be logged and provided with the purpose of behavior analysis for improving quality of life (see Fig. \ref{figure:overall}).

\begin{figure}[t]
\begin{center}
\includegraphics[width=0.6\linewidth]{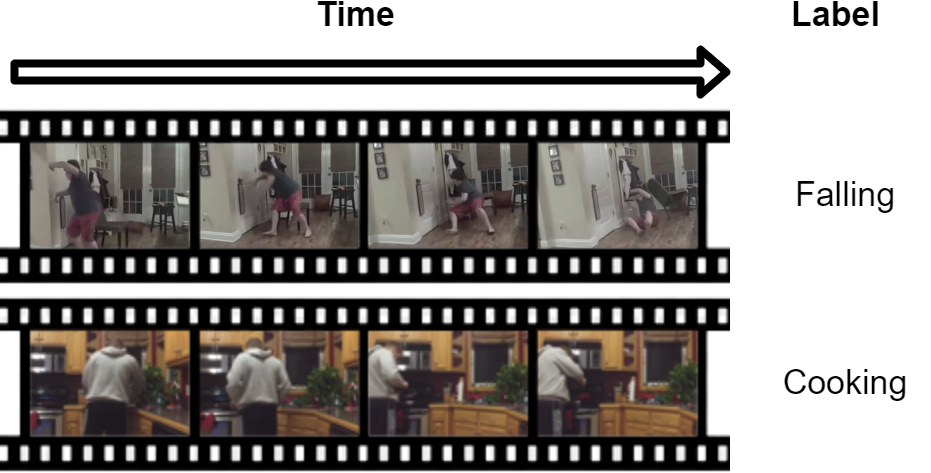}
\end{center}
    \caption{Example of a sequence of frames representing two actions: (top) a person falling down, (bottom) a person cooking.}
\label{figure:sequence-actions}
\end{figure}

Deep learning (DL) models are based on large architectures of multi-layer neural networks. In recent years, DL have greatly advanced the state of the art in Computer Vision. We based our activity recognition module on DL models, due to its performance in comparison with approaches based on hand-crafted features \cite{yao2019review}. We design and train our model for indoor monitoring action recognition using a technique called Transfer Learning. It is used when the available examples for training Machine Learning methods are not enough \cite{thrun1998learning}. Transfer Learning enables improvements in the recognition performance specializing networks that have been previously trained on large datasets that cover thousands of common daily actions \cite{carreira2017quo}. In order to enable local edge processing, the approach is embedded in Nvidia Jetson TX2 devices, and performance is maximized to reach real-time processing. Finally, local processing in embedded SoCs greatly reduces data bandwidth and prevents the transmission of raw images. This is a key point to preserve the privacy of sensitive data that we are processing in our indoor monitoring system.

Regarding online monitoring, we use a remote platform that collects information about the detected actions, and qualities of the local edge nodes such as energy consumption, temperature, or processing time performance (in frames per second, FPS). Data is analyzed by the QRM tool that makes decisions about it, in order to optimize parameters for the overall system. In this work, we are showing an example of energy-efficient CPS reconfiguration. After reaching some energy consumption threshold, the QRM tool sends a command to the action recognizer to switch the DL-based model in order to maximize the working time of the device (for battery-powered devices). Although reconfiguration is triggered by energy, also the overall average recognition precision of the system vs processing time trade-offs are impacted. Finally, the flexibility of our versatile CPS proposal allows us to perform other reconfigurations.

Our main contributions in this work are: 1) the design and development of different DL-based Human Action Recognizer modules for indoor monitoring using Transfer Learning, and the collection of an indoor monitoring dataset; 2) the implementation of the indoor monitoring module embedded on local node SoCs (Nvidia Jetson TX2): it achieves real-time processing, reduces data bandwidth, and preserves privacy of sensitive data; 3) the development of tools for online monitoring of energy consumption of the local edge nodes in a distributed scenario; 4) the design and integration of a reconfigurable CPS for indoor monitoring integrating the HAR module, and a remote platform for online resource monitoring and quality and resource management. The final reconfigurable CPS is energy aware, adapting the architecture to maximizing working time or minimizing energy consumption.

The rest of the paper is organized as follows: in \S\ref{sec:related}, we review the related methods for reconfigurable CPS, life-style monitoring and DL architectures for HAR. In \S\ref{sec:reconfigurable}, we describe our DL models and the properties of our QRM tool. In \S\ref{sec:results} we compare the developed DL architectures in terms of accuracy, and other qualities such as processing time and power consumption. Finally, we study the system operating time depending on whether a reconfiguration is triggered or not.

\section{Related methods}
\label{sec:related}

\subsection{Reconfigurable Cyber-Physical Systems}
\label{subsec:reconf-cps}
In recent years, novel reconfigurable CPSs have been proposed to self-adapt to their changing environment \cite{boschi2016functional}. Reconfigurable CPSs not only offer flexibility in terms of functionality, but also aim at accomplishing the optimization of certain qualities when carrying out a task. However, systems provide different levels of autonomy, with reconfiguration triggered by a user \cite{nogueira2019self} or automatically triggered by the system when monitoring environment conditions or performance are met \cite{gaur2019cap}. For example, in \cite{brusaferri2014cps} authors propose a reconfigurable CPS for manufacturing plants using Virtual Avatars that automatically adapt to new automation architectures and performance or \cite{bonci2019cyber} for efficiency monitoring, integrating overall throughput effectiveness approaches via automatic reconfiguration. Contrary to automatic adaptation, some systems show that human intervention is crucial, such as in \cite{nogueira2019self} that proposes a CPS for collaborative robotics in automotive industry built around a strong human-robot interaction.

\subsection{Life-style monitoring systems}
\label{subsec:lf-monitor-sys}
The main goal of lifestyle monitoring systems is to analyze actions being carried out, specially for the elderly or people that require supervision \cite{er2018non}. Recently, Human Action Recognition  became popular and HAR systems can be divided into two main categories: 1) wearable sensors HAR systems, analyzing activities from smartphones \cite{zeng2014convolutional}, or accelerometers \cite{chen2015deep}; 2) video-based HAR systems that analyze a sequence of frames (see Fig. \ref{figure:sequence-actions}) to recognize the action \cite{choutas2018potion}. In our approach, we use video-based method powered by DL architectures that facilitate accurate recognition, and are less invasive since they do not require to wear any device. Bear in mind that for example, the elderly are usually reluctant to wear such devices or to use mobile apps.

\begin{table}[t]
\centering
\caption{Generic action recognition datasets, number of action classes and samples, and actions used in our custom dataset}
\label{table:1}
\begin{tabular}{ | c | c | c || c|}
\hline
Dataset                & \#Actions & \#Samples & \#Actions used \\
\hline
HMDB51  \cite{kuehne2011hmdb}  & 51      & 6766  & 10\\
UCF-101 \cite{soomro2012ucf101} & 101     & 13320  & 2\\
Fall Detection \cite{charfi2013optimized} & 2       & 222 &    1\\
Charades \cite{sigurdsson2016hollywood} & 157     & 66500  & 5\\
STAIR \cite{yoshikawa2018stair}  & 100     & 102462 & 5\\
Kinetics \cite{kay2017kinetics}  & 600     & 495547 & 16\\
\hline
\end{tabular}
\end{table}

\subsection{Deep Learning architectures for HAR}
\label{subsec:DL}

Advanced Machine learning techniques based on Deep Learning architectures are the state of the art for monitoring. The architectures used in this work include:
\begin{itemize}
    \item 3D Convolutional Networks (3D CNN) for video \cite{tran2015learning}. 3D CNN layers extract spatio-temporal features from the sequences of frames. The work in \cite{tran2015learning} outperforms the other state-of-the-art methods in terms of accuracy. However, 3D convolutions lead to a large increase in the number of parameters compared to 2D convolutions. Thus, a large amount of training data is required.
    \item R(2+1)D Networks \cite{tran2018closer}. The model proposes a factorization of 3D convolutions into 2D+1D convolutions to reduce the computational complexity. In this way, 2D convolutions extract spatial information, and 1D convolutions extract temporal features. This approach obtains gains in terms of accuracy due to the increase of non linearities.
    \item Recurrent Convolutional Networks (RCN) \cite{donahue2015long}. They are based on the use of two-dimensional convolutional neural networks (2D CNN) in combination with Long Short-Term Memory (LSTM) units or Gated Recurrent Units (GRU) layers \cite{cho2014learning}. Frames are processed individually by a 2D CNN to automatically extract spatial features that are then fed into the recurrent units, that learn sequences of these features.
\end{itemize}

As mentioned in the Introduction, our indoor monitoring system uses RGB video streams, and an essential aspect in this research field is the availability of training data and annotated datasets for benchmarking. In our work, we have selected some of the available datasets in the state of the art: HMDB \cite{kuehne2011hmdb}, UCF-101 \cite{soomro2012ucf101}, Charades \cite{sigurdsson2016hollywood}, and Kinetics \cite{kay2017kinetics} (see Table \ref{table:1}). Our purpose is to use a pre-trained network architecture in a general action dataset such as Kinetics. After that, we will apply Transfer Learning to specialize our network, partially re-training the network on actions selected from all the datasets. This considerably reduces training time while still guarantees generalization \cite{carreira2017quo}. Our selection criteria is indoor actions that provide information about lifestyle.

\begin{figure}[t]
\centering
\includegraphics[width=0.60\linewidth]{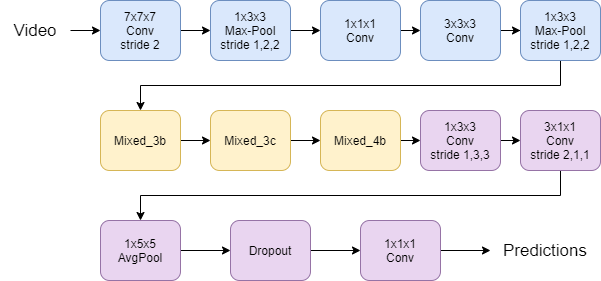}
\caption{\textit{RGBI3D reduced} model architecture. Blue and orange layers are part of the original model (weights after training on Kinetics \cite{carreira2017quo}). In \textit{RGBI3D reduced}, purple layers substitute the top layers of \textit{RGBI3D original} and are re-trained using our custom dataset for indoor lifestyle monitoring.}
\label{figure:reducedi3d}
\end{figure}

\section{Reconfigurable CPS for indoor monitoring}
\label{sec:reconfigurable}

Our proposal is organized around two main components: 1) the action recognizer module, based on different architectures using DL techniques attending to qualities such as average accuracy, processing time performance (in fps), power consumption, and size of the model mainly determined by the number of parameters; 2) the design of an adaptable and reconfigurable system by using a remote online monitoring platform and a QRM tool, that is capable of triggering reconfiguration at local edges based on the analysis of the logged data.

\subsection{CPS core: Local HAR and global data collection}
\label{subsec:cpscore}

Regarding the collection of data, HAR is embedded in local edges. A separate module gathers all the processed information from the different edge nodes, and sends it to the online monitoring tool. 

The HAR automatically classifies a sequence of frames labeling it as an indoor action. For the implementation of this module we take into account three aspects: gathering the custom dataset to train our models for lifestyle indoor monitoring, the network architecture for the Deep Neural Network model, and the training process.

\subsection*{Custom dataset for lifestyle indoor monitoring}
\label{subsubsec:dataset}

Although current datasets available in the state of the art provide thousands of actions, none of the datasets mentioned in \S\ref{subsec:DL} includes enough relevant actions for lifestyle indoor monitoring. Therefore, the classification metrics for these specific actions are very poor. 

To overcome this problem, we collected video clips from different datasets, choosing 18 actions that we considered relevant (see Table \ref{table:dataset_stats}). Since some actions are under-represented, manually-labeled clips from YouTube videos were cropped and added. We also included a "no action" class, combining the following two approaches: 1) video examples generated from the Indoor Scene Recognition dataset \cite{quattoni2009recognizing} that include indoor empty spaces; 2) video examples from YouTube virtual house tours that show house walkthroughs without people (about 950 clips).

Our custom dataset for lifestyle monitoring gathers 18 actions of interest (clips limited to 10 seconds). Also, the dataset is highly imbalanced: some classes with more than 1500 clips, while others have less than 500. This is a well-known common issue in DL; an imbalanced training set leads to models that tend to learn better actions with higher number of clips. Consequently, the model disregards classes with few examples. Our solution takes this into account and carries out the optimization weighing the loss value at training, using the number of examples per class. 

\begin{table}[t]
    \begin{minipage}[t]{.5\linewidth}
    \caption{Our dataset: number of clips retrieved per class}
    \label{table:dataset_stats}
        \begin{tabular}{| c | c | c | c |}
        \hline
        Action Classes & \# Train & \# Validation & \# Test \\ 
        \hline
        bandaging  & 619 & 48  & 99  \\
        blowing nose or sneezing & 1140 & 99 & 187 \\
        cleaning floor & 1876 & 160 & 298 \\
        cooking & 1965 & 186 & 332 \\
        eating  & 1183 & 165 & 215 \\
        falling down & 808 & 72 & 121 \\
        hitting  & 1014 & 74 & 120 \\
        lying on bed or sofa   & 300 & 39 & 40 \\
        lying on the floor  & 1186 & 148 & 148 \\
        running & 403 & 48 & 49 \\
        sitting & 857 & 103 & 104 \\
        sleeping & 574 & 47 & 96 \\
        smoking  & 895 & 54 & 81 \\
        standing up & 549 & 65 & 66 \\
        using inhaler  & 533 & 46 & 96 \\
        walking & 1212 & 147 & 148 \\
        watching tv & 541 & 48 & 95 \\
        no action & 1500 & 200 & 200 \\
        \hline
        Total   & 17155 & 1749 & 2495 \\
        \hline                                                                         
        \end{tabular}
    \end{minipage}\qquad
    \begin{minipage}[t]{.5\linewidth}
        \caption{Local edge operating modes}
        \label{table:configuration-states}
        \begin{tabular}{| c | c | c | c | c | c|}
        \hline
        Mode & DL model & Size & GPU & Device \\
        id &  & (MB) & (W) & (W) \\
        \hline
        0        & \textit{RGBI3D original}        & 49.9   & 1.58   & 4.77 \\
        1        & \textit{RGBI3D reduced} & 47.6   & 1.37   & 4.43  \\
        2        & \textit{RGB2D Mobilenet GRU} & 26.9   & 0.25   & 2.61 \\
        \hline
        \end{tabular}
    \end{minipage}
\end{table}

\subsection*{Deep Neural Network architecture}
\label{subsubsec:network}

We choose the RGBI3D architecture as the basis for our network models. The model parameters or weights for the different layers are publicly available after being trained on the large-scale action video dataset Kinetics600 \cite{carreira2018short}. This allows us training an indoor HAR model in an affordable amount of time on a single GPU (RTX 2080 Ti). Then, we apply Transfer Learning for specializing the architecture for our custom dataset.

In this work, we implement four alternatives for HAR (two of them based on the RGBI3D model mentioned above):
\begin{itemize}
    \item \textit{RGBI3D original}: We re-train the top layers of the original RGBI3D architecture on our custom dataset. The base model layers are frozen and the model is trained only for 26 epochs. This stream extracts spatio-temporal information from the video. The architecture has more than 12 million parameters (49.9 MB).
    \item \textit{RGBI3D reduced}: Taking into account that the HAR is to be embedded in a device with a limited amount of resources, we propose a simpler model that will require fewer resources. However, accuracy is also reduced. This model has two parts: 1) the RGBI3D model up till the layer \textit{Mixed\_4b}, using the pre-trained weights of this basis to extract general features from the input videos; 2) For the remaining layers, we followed an approach inspired by \cite{tran2018closer}. We include factorized 3D convolutions, simplified as 2D plus 1D convolutions (2+1)D. After the (2+1)D convolution layers, we use the top layers from the original model. The resulting model has only 5.5 million parameters (less than half of the original), and 47.6 MB. It is trained for 45 epochs. The \textit{RGBI3D reduced} architecture is shown in Fig. \ref{figure:reducedi3d}. 
    \item \textit{RGB2D Mobilenet GRU}: This architecture is inspired in \cite{donahue2015long}. 5 equidistant frames are selected out of the batch of 64. Every frame is passed to the Mobilenet \cite{howard2017mobilenets} network in order to extract spatial features. These features are then passed through a GRU layer. This layer aims at learning temporal dynamics of the actions. This architecture has only 4.38 million parameters, with a size in MB 40\% smaller than \textit{RGBI3D original}, and it is trained for 16 epochs. Bear in mind that Mobilenet was originally developed to improve the real-time performance of Deep Learning for platforms with limited hardware resources.
    \item \textit{RGB3D fully trained}: This is a simplified version of \textit{RGBI3D reduced} for which no Transfer Learning is applied. The architecture is completely trained from randomly initialized weights. Due to memory limitations, only 32 non-consecutive frames are selected to train the network. Also, the training is done for 37 epochs. The final model reaches 2.6 million parameters (31.7 MB). 
\end{itemize}

Regarding the implementation, we use batches of 64 frames as our input. Considering that every video is recorded at 25 fps, the network analyzes 2.5~s at every step. We perform data augmentation on the training stage to prevent the model from overfitting. The video resolution is downsampled until each frame short side is equal to 256 pixels. Then a random window of 224~x~224 pixels of 64 frames is selected from the training instance, which is the network input size. Additional augmentations include random horizontal flip, and 2D rotations of 0 to 5 degrees. 

We use the Keras framework to design and train the models. With respect to the parameters, we use the Adam optimizer with a starting learning rate of 0.0001.

\subsection{Monitoring and reconfiguration}
\label{subsec:adaptable}

The second part of our reconfigurable CPS is a remote online monitoring platform and a QRM tool. The HAR module is performed every 64 frames, gathering information about the recognized actions, and hardware qualities such as power consumption, temperature, and performance from the different local nodes. All the information is continuously sent to the monitoring platform that analyzes the data.

In this work, we are presenting an energy-efficient CPS that selects the DL model to be automatically deployed to the local edge nodes. The proposed DL models (in \S\ref{subsubsec:network}) provide our lifestyle indoor monitoring CPS several alternatives with different power vs. accuracy trade-offs. Therefore, our energy-efficient CPS can be also reconfigured adapting it to different scenarios and re-deploy a different DL model in the local edges. 

Specifically, the HAR module is able to operate in three different modes shown in Table \ref{table:configuration-states}. \textit{Mode 0} uses the most resource-demanding DL model, that also achieves the highest accuracy when testing on our custom dataset. Secondly, \textit{mode 1} uses the \textit{RGBI3D reduced} DL model that is a simplified version of \textit{RGBI3D original} with lower energy consumption. Finally, \textit{mode 2} uses the \textit{RGB2D Mobilenet GRU} DL model that achieves the highest performance in fps, considerably reducing the energy consumption. In every mode, the operating frequency of the hardware components of the local edge devices are set to its minimum value in order to reduce the power consumption of the HAR module, and thus maximizing battery lives.

The quality and resource management tool is the module that decides whether reconfiguration is triggered. Table \ref{table:policies} shows reconfiguration scenarios with different policies that are analyzed in \S\ref{sec:results}. The QRM tool drives reconfiguration by monitoring the average energy consumption and the remaining battery percentage, re-deploying the fittest DL model to the local edges.

Moreover, the analysis of the monitored data is meant to be stored and presented in a lifestyle habit summary that includes valuable information about: sedentariness, physical exercise, meal routines, or medical control and supervision.  
 
\begin{table}[t]
\centering
\caption{Evaluation of metrics for the proposed models}
\label{table:eval-metrics}
\begin{tabular}{| c | c | c | c | c |}
\hline
DL Model & Accuracy & Precision & Recall & F-1 score \\
\hline
\textit{RGBI3D original}         & 84.24  & 77.83    & 79.30 & 78.14 \\
\textit{RGBI3D reduced} &  81.00 & 76.65    & 75.81 & 74.99 \\
\textit{RGB2D Mobilenet GRU} & 76.27  & 69.99 & 70.95 & 70.08 \\
\textit{RGB3D fully trained} & 65.69 & 58.77 & 60.26 & 58.35 \\
\hline
\end{tabular}
\end{table} 

\begin{table}[t]
\caption{Definition of policies for the reconfiguration scenarios}
\label{table:policies}
\centering
\begin{tabular}{| c | c | c | c | }
\hline
 Policies &\multicolumn{3}{ | c |}{Remaining Battery Percentage} \\ 
\hline
- & 100-51\% & 50-26\% & 25-0\% \\
\hline
Scenario 1 &	Mode 0	&   Mode 0	&   Mode 0 \\
Scenario 2 &	Mode 1 &	Mode 1 &	Mode 1 \\
Scenario 3 &	Mode 2 &	Mode 2 &	Mode 2 \\
\hline
Scenario 4 &	Mode 0 &	Mode 1 &	Mode 1 \\
Scenario 5 &	Mode 0 &	Mode 2 &	Mode 2 \\
Scenario 6 &	Mode 1 &	Mode 2 &	Mode 2 \\
Scenario 7 &	Mode 0 &	Mode 1 &	Mode 2 \\
\hline
\end{tabular}
\end{table}

\begin{figure}[t]
\begin{center}
\includegraphics[width=0.45\linewidth]{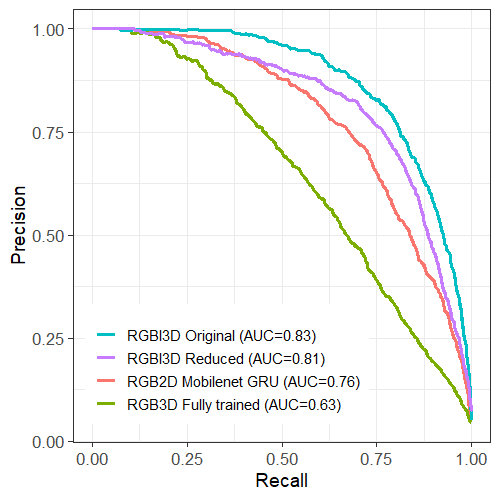}
\end{center}
\caption{Macro-Average Precision-Recall Curve for the considered DL models. \textit{RGBI3D original} achieves the larger area under the curve, thus reaching high precision and recall values. Note how \textit{RGBI3D reduced} achieves very similar results while considerably simplifying the model, while \textit{RGBI3D fully trained} performs poorly.}
\label{figure:pr-curve}
\end{figure}

\section{Results and discussion}
\label{sec:results}

\begin{figure*}[t]
    \centering
    \includegraphics[width=0.7\textwidth]{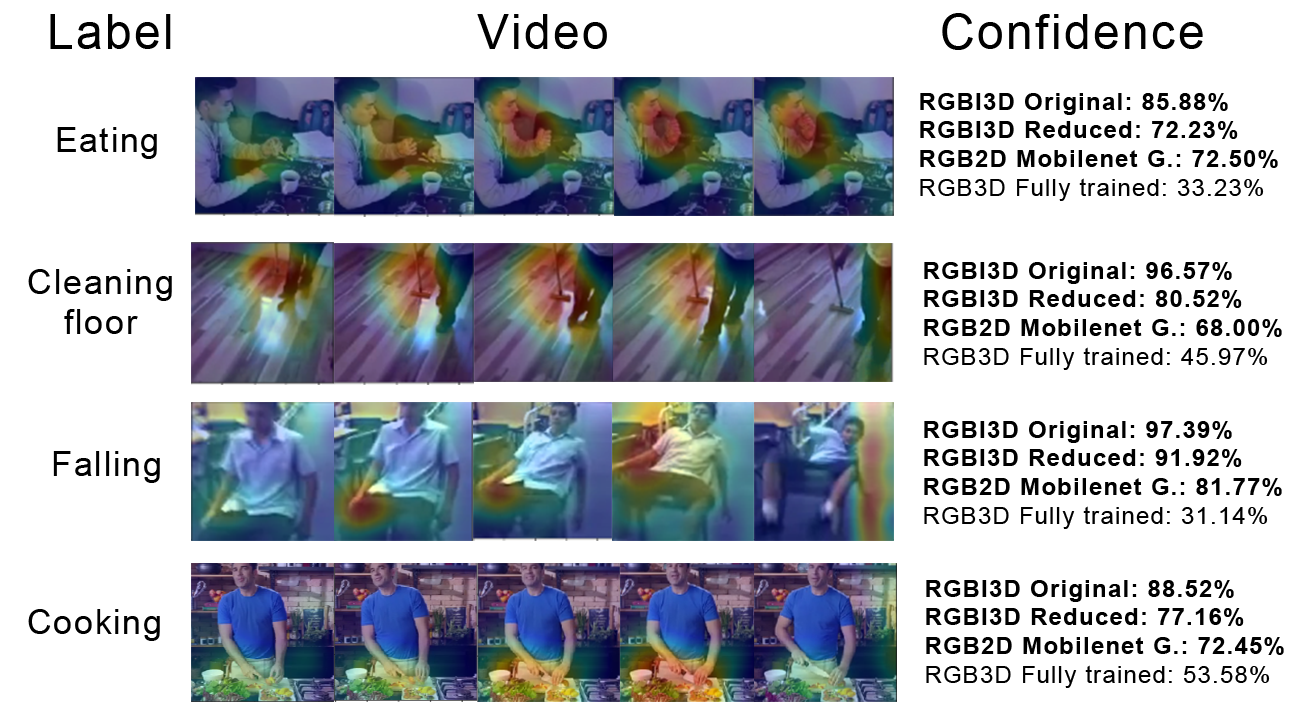}
    \caption{Qualitative examples for some videos only used for testing from our lifestyle indoor monitoring dataset. For each example, we show the ground-truth label, a sequence of frames with overlapping heat maps, and the confidence labels for each DL model. Heat maps are computed using the \textit{RGBI3D original}, the model with the highest confidence. Heat maps show the focus regions where the DL model extracts the features for action recognition. For example, in \textit{cooking} and \textit{eating} videos, hands provide most of valuable information for recognition while in \textit{cleaning floor} the attention is focused on the sweeper.}
    \label{figure:actions-grad-cam}
\end{figure*}

\begin{figure*}[t]
    \begin{center}
        \begin{minipage}[t]{0.33\textwidth}
	        \centering
 	        \includegraphics[width=\textwidth]{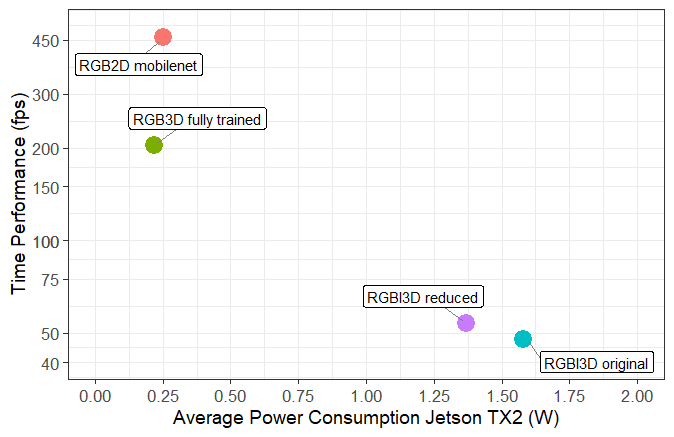}
        \end{minipage}%
        \begin{minipage}[t]{0.33\textwidth}
	        \centering
 	        \includegraphics[width=\textwidth]{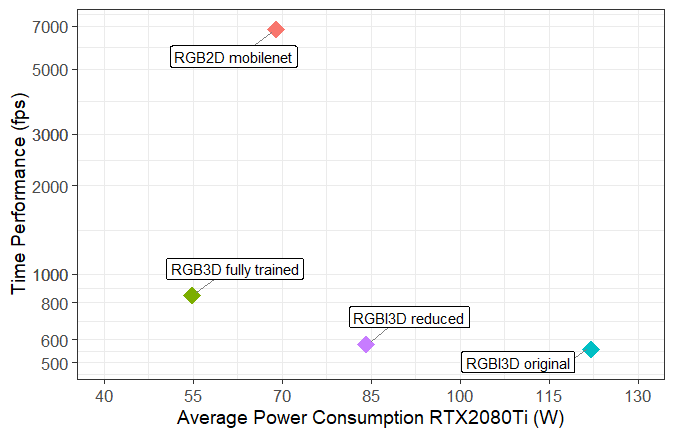}
        \end{minipage}%
        \begin{minipage}[t]{0.33\textwidth}
	    \centering
 	        \includegraphics[width=\textwidth]{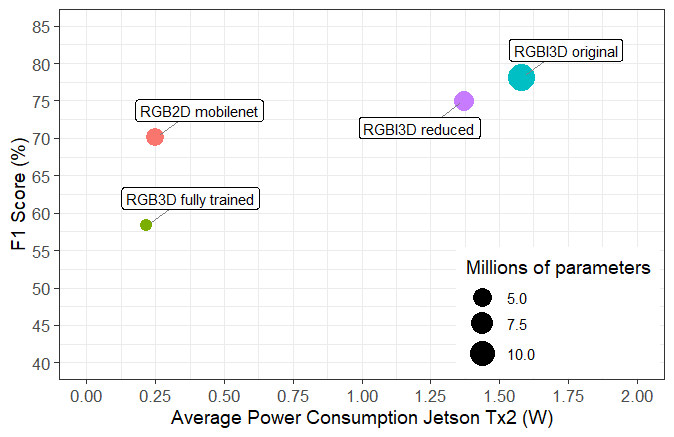}
        \end{minipage}
    \end{center}
    \caption{Left and center: Time performance (fps) Vs Power Consumption for Jetson TX2 and RTX2080Ti platforms respectively. Right: Macro-F1 Score Vs Power Consumption on Jetson TX2 and size (million  parameters)}
    \label{figure:performance-comparison}
\end{figure*}

In this section, we analyze the accuracy of the proposed DL models, the energy consumption compared to processing time performance, the size of our model implementations, and the energy vs. working time trade-offs for the reconfiguration policies.

Regarding accuracy, every DL model was evaluated using our custom dataset for lifestyle indoor monitoring. Videos were clipped in batches of 64 frames, and then inference confidence was averaged over all the batches. As mentioned in \S\ref{subsubsec:dataset}, our dataset for multiclass classification is highly imbalanced: metrics to evaluate the DL models are adapted, including precision ($\frac{TP}{TP+FP}$), recall ($\frac{TP}{TP+FN}$), and F1 score (harmonic mean of precision and recall). For the multiclass recognition, macro-metrics that are arithmetic means of the per-class metrics are computed, in order to equally weigh every class. For the sake of simplicity, from here we refer to macro-metrics just as precision, recall, and F1 score. Table \ref{table:eval-metrics} presents the evaluation of the DL models using these metrics. Results show that \textit{RGBI3D original} outperforms the rest of the DL models, not only in terms of accuracy, but also in terms of the F1 score. This means that this model is the one that correctly classifies, in average, most of the classes. In spite of this, the F1 score is lower than the accuracy, and this is something common in our approaches. This is because our architectures correctly classify classes with lots of examples, but are prone to fail when training examples for some classes are not enough. In order to overcome this problem, we also compute the macro F1 score to compare the performance of each DL model, since accuracy does not take into account the effect of imbalanced datasets. 

Finally, Fig. \ref{figure:pr-curve} compares the precision-recall curves for each DL model. PR-curves show the trade-off between the precision and recall metrics for different thresholds. Larger values of area under the curve (AUC) denote high precision and recall values. Obviously, the original and more complex model achieves the best values. However, these values are very similar for the simplified \textit{RGBI3D reduced} model with a drop of only 2\% AUC. Contrarily, the \textit{RGB3D fully trained} model performance is significantly worse. Consequently, this model is not included in our choices for CPS reconfiguration. This fact supports the potential of Transfer Learning when dealing with not enough examples in our dataset. \textit{RGB3D fully trained} is included here to offer a general comparison and show the benefit of DL models pre-trained in large general datasets. 

In Fig. \ref{figure:actions-grad-cam} we illustrate the HAR performance with some qualitative results. On the right column we report the model confidence for these test examples, for which \textit{RGBI3D original} achieves the best results, closely followed by \textit{RGBI3D reduced}. Grad-Cam \cite{selvaraju2017grad} is used to generate the visual explanation of the areas where \textit{RGBI3D original} automatically focuses to predict the label. This is represented with overlapped heat maps. Note that the recognition focuses on the area around the hands for  \textit{eating}, or tools for  \textit{cleaning}.

Regarding the hardware devices, our HAR module is designed to be embedded in an Nvidia Jetson TX2, taking into account that the DL models fit given the device GPU memory. The Jetson TX2 module is a SoC with a six-core CPU with 8GB DDR4 memory and a GPU with 256 CUDA cores. 

With respect to the energy efficiency, we measure the average power consumption for each alternative in Watts (W) performing 30 inferences, and recording the current energy consumption of the GPU every 100 ms. Fig. \ref{figure:performance-comparison} presents a comparison of our DL models implemented in the local edges (left: Nvidia Jetson Tx2) versus a high-performance GPU (center: Nvidia RTX 2080Ti). Firstly, note that the significant difference in average power consumption with 20x - 30x reduction for our SoC platform, that is more power-efficient thanks to the hardware architecture improvements, its operating frequency, and available resources. Secondly, more complex DL models require more resources and reducing the processing time. For example, while \textit{RGBI3D original} and \textit{RGBI3D reduced} process batches of 64 frames, the simpler \textit{RGB2D Mobilenet GRU} processes batches of 5 frames. Obviously, without attending to the number of layers or their size, only this difference in the input represents a significant reduction in the number of operations to be carried out.

Every DL model receives an input video at 25 fps, thus processing time performance (fps) greater than that enables our DL models to operate in a multi-view environment, ensuring real-time performance. All this explains \textit{RGBI3D original} processing time: it is about 10x slower than the simpler \textit{RGB2D Mobilenet GRU} while requires almost the double energy, for both the Nvidia Jetson and the Nvidia RTX implementations. Finally, there is a very small difference in average power consumption between the \textit{RGBI3D original} and the \textit{RGBI3D reduced} models on the Jetson TX2, but it represents around 15-20\% reduction. For the RTX platform, the reduction in power consumption reaches 30\%. Finally, note also the Jetson TX2 consumption is around 1/75th of the power consumption of the RTX platform (98\% reduction).

Fig. \ref{figure:performance-comparison}-right shows additional facets for comparison, specifically the F1 score and power consumption for the Jetson TX2 local edge computing nodes, and DL model sizes in million parameters. First, models with higher F1 score also require more resources and thus, more power consumption and memory. In other words, models with more parameters reach better classification at the expense of energy cost. In our case, we need a good accuracy vs energy consumption trade-off, facing reconfiguration to re-deploy DL models on local edges. Different scenarios are analyzed in Table \ref{table:policies-eval}, with the global aim of extending battery-powered working time while preserving accuracy. 

Taking into account the power consumption for 10 hours of working time for \textit{Scenario 1}, we estimated the overall working time for the other scenarios in Table \ref{table:policies}. Note that this working time represents continuous operation; in a real-world scenario this represents weeks of operation because the HAR module does not operate if no activity is detected. For the first three scenarios, only using one of the models for the whole time, working time varies from 10 (\textit{Scenario 1}) to more than 18 hours (\textit{Scenario 3}) losing 10\% of F1 score in that range. For the next 4 scenarios, the reconfiguration is triggered to change the DL-model, achieving different working time vs accuracy trade-offs within these ranges. The calculated F1 score is a weighted average of F1 score over the working time, taking into account the values estimated in Table \ref{table:eval-metrics} for each DL model alternative. In general, \textit{Scenarios 4 to 7} for which reconfiguration is performed, provide significant improvements in operating time when compared to \textit{Scenarios 1 to 3} where no changes happen. Finally, in \textit{Scenario 7} reconfiguration is triggered twice extending around 22\% the operating time (with respect to \textit{Scenario 1}) and limiting the F1-score loss to only 3.7 points.

Regarding scalability, new computing nodes can be added to the network without signiﬁcant changes in the system architecture, except for monitoring and analysis of the new data. This represents a great beneﬁt for nursing homes, hospitals, or healthcare public administrations: monitoring the health of the elderly while encouraging them to be more active, or helping caretakers check the elderly daily routines such as taking their pills, meals, etc. Moreover, local processing also reduces data transmission. This is important at a home level with bandwidth constraints and also associated energy costs. Optimizing bandwidth consumption and energy cost is key for a low invasive service, without requiring users to adapt their bandwidth and energy provider contracts. The capability of remote reconfiguration of the platform allows better customization of the service for different service profiles and thus optimize the benefit vs cost trade-off.

\begin{table}[t]
\caption{Trade-off between F1 score and device working time for the reconfiguration policies}
\label{table:policies-eval}
\centering
\begin{tabular}{| c | c | c | c | }
\hline
 Policies   & Reconfiguration & Working time & Average F1 \\ 
\hline
\textit{Scenario 1} & off & 10h 00' & 78.14 \\
\textit{Scenario 2} & off &	10h 46' &	74.99 \\
\textit{Scenario 3} & off &	18h 18' &	70.08 \\
\hline
\textit{Scenario 4} & on & 10h 23' &	76.51 \\
\textit{Scenario 5} & on & 14h 09' &	72.93 \\
\textit{Scenario 6} & on &	14h 32' &	71.90 \\
\textit{Scenario 7} & on &	12h 16' &	74.44 \\
\hline
\end{tabular}
\end{table}

\section{Conclusions}
\label{sec:conclusion}
This work proposes an energy-aware CPS for lifestyle indoor monitoring, optimizing the overall working time vs accuracy trade-off of the distributed system via reconfiguration. For the physical interaction with the real world, we propose embedded HAR analysis using Jetson TX2. Our local edges for HAR are built on DL architectures that achieve high F1 scores consuming only a few Watts. For this, DL models that reuse pre-trained weights and are afterwards specialized, have been shown very valuable. Also, real-time video processing enables rapid adaptation of our system closely monitoring the considered qualities. Finally, note the relevance of local processing for the preservation of privacy. In our system, no images are transmitted to the network, only label confidence and qualities. Thus, this approach inherently protects sensitive user data. 

Reconfiguration makes CPS more flexible and allows us to add new functionalities. In our case, by switching the active DL models running on the local edge nodes, the device total working time can be extended for up to 20-30\% depending on the policy, compared to the scenarios where no reconfiguration is considered. At the same time, accuracy is preserved losing only 5\% of the average weighted F1 score along the extended working time.

In future work, we plan new reconfiguration policies based on different qualities such as the processing time and the confidence of recognized action labels to resolve ambiguous scenes. Also, new frameworks for computational load balance, low-latency network topologies, and sensitive data protection will be studied.

\section*{Acknowledgement}
This work was partially supported by the FitOptiVis project funded by the ECSEL Joint Undertaking, grant H2020-ECSEL-2017-2-783162, and the Spanish National grant funded by MINECO through APCIN PCI2018-093184.

\bibliographystyle{unsrt}  
\bibliography{manuscript}

\end{document}